# Title: Nurse Rostering with Genetic Algorithms

Uwe Aickelin
School of Computer Science
University of Nottingham
NG8 1BB   UK
uxa@cs.nott.ac.uk

Affiliations:    OR Society

In recent years genetic algorithms have emerged as a useful tool for the heuristic solution of complex discrete optimisation problems. In particular there has been considerable interest in their use in tackling problems arising in the areas of scheduling and timetabling. However, the classical genetic algorithm paradigm is not well equipped to handle constraints and successful implementations usually require some sort of modification to enable the search to exploit problem specific knowledge in order to overcome this shortcoming. This paper is concerned with the development of a family of genetic algorithms for the solution of a nurse rostering problem at a major UK hospital.

The hospital is made up of wards of up to 30 nurses. Each ward has its own group of nurses whose shifts have to be scheduled on a weekly basis. In addition to fulfilling the minimum demand for staff over three daily shifts, nurses' wishes and qualifications have to be taken into account. The schedules must also be seen to be fair, in that unpopular shifts have to be spread evenly amongst all nurses, and other restrictions, such as team nursing and special conditions for senior staff, have to be satisfied.

The basis of the family of genetic algorithms is a classical genetic algorithm consisting of n-point crossover, single-bit mutation and a rank-based selection. The solution space consists of all schedules in which each nurse works the required number of shifts, but the remaining constraints, both hard and soft, are relaxed and penalised in the fitness function.

The talk will start with a detailed description of the problem and the initial implementation and will go on to highlight the shortcomings of such an approach, in terms of the key element of balancing feasibility, i.e. covering the demand and work regulations, and quality, as measured by the nurses' preferences. A series of experiments involving parameter adaptation, niching, intelligent weights, delta coding, local hill climbing, migration and special selection rules will then be outlined and it will be shown how a series of these enhancements were able to eradicate these difficulties.

Results based on several months' real data will be used to measure the impact of each modification, and to show that the final algorithm is able to compete with a tabu search approach currently employed at the hospital. The talk will conclude with some observations as to the overall quality of this approach to this and similar problems.

# Nurse Rostering with Genetic Algorithms

by

Uwe Aickelin

# Overview

- Introduction to the Nurse Scheduling Problem and corresponding IP.

- Introduction to the standard genetic algorithm.

- Application of the genetic algorithm to the nurse scheduling problem.

- Limitations of the standard approach.

- Enhancements of the standard approach.

- Final results and comparison.

- Future scope of research.

- References.

# The Nurse Scheduling Problem

## Objective:

- To create weekly schedules on ward basis.

- To satisfy working contracts and to have fair schedules.

- To take as many nurses' requests into account as possible.

## Decomposition:

1. Ensuring that nurses present can cover the overall demand.

2. Scheduling the days and/or nights a nurse works.

3. Splitting the day shifts into early and late shifts.

## Typical Dimensions of Data:

30 nurses, 3 grade bands, 7 part time options, 411 different shift patterns, varying demand levels.

# IP Model

$$x_{ij} = \begin{cases} 1 & \text{nurse i works pattern j} \\ 0 & \text{else} \end{cases}$$

$$a_{jk} = \begin{cases} 1 & \text{pattern j covers day k} \\ 0 & \text{else} \end{cases}$$

$$q_{is} = \begin{cases} 1 & \text{nurse i is of grade s or higher} \\ 0 & \text{else} \end{cases}$$

$p_{ij}$ = penalty cost of nurse i working pattern j
$R_{ks}$ = demand of nurses with grade s on day k
$F(i)$ = set of feasible shift patterns for nurse i

**Target Function:**

$$\sum_{i=1}^{n} \sum_{j=1}^{m} p_{ij} x_{ij} \rightarrow \min!$$

**1. Everybody works exactly one pattern:**

$$\sum_{j \in F(i)} x_{ij} = 1 \qquad \forall i$$

**2. The demand is covered for every grade on every day:**

$$\sum_{i \in F(i)} \sum_{j=1}^{m} q_{is} a_{jk} x_{ij} \geq R_{ks} \qquad \forall k, s$$

# Genetic Algorithms (GAs)

- A Heuristic based on the principles of natural evolution and 'survival of the fittest'.

**Population:** The GA works with many solutions at the same time. New Solutions inherit good parts from old solutions.

**Coding:** Transformation of variables into chromosomes such that genetic operators can be applied to solutions.

**Fitness:** The fitter a solution, the more likely it will contribute to new solutions. Based on the target function.

**Selection:** Rank-based: Individuals (solutions) are ranked according to fitness. The higher the rank the more likely an individual is chosen as a parent.

**Crossover:** Combining parts of parent individuals (cut and paste) to create new solutions: 'Building Block Hypotheses'.

**Mutation:** Random change of a single bit of an individual.

**Replacement:** 'Elitist' Strategy, i.e. the best X% of the old solutions are kept. The rest are replaced.

# Application of the GA to the Scheduling Problem

**<u>Proposed Coding:</u>**

- Each individual represents a full one week schedule.

- It is a string of n elements, n being the number of nurses.

- Each element is the index of the shift pattern worked by a nurse.

➔ Crossover gives some nurses the shifts worked in one parent solution and the remainder those worked in the other parent.

➔ Mutation changes the worked shift of one nurse.

**<u>Dealing with constraints:</u>**

- Implement constraints into the coding.

- Design a repair operator.

- Add a penalty to an infeasible solution's objective function value.

### New target function:

$$\sum_{i=1}^{n}\sum_{j=1}^{m} p_{ij} x_{ij} + g_{demand} \sum_{k=1}^{14}\sum_{s=1}^{p} \min\left[\left|\left(\sum_{i=1}^{n}\sum_{j=1}^{m} q_{is} a_{jk} x_{ij}\right) - R_{ks}\; ;\; 0\right|^{+}\right] \rightarrow \min!$$

→ 28.6% feasible solutions. Average objective value of the best feasible solutions 38.2 (Branch and Bound ≈ 11).

# Parameter Optimisation and Dynamic Weights

**Parameters to be set:**

- Population size (100 – 2000).

- Type of crossover (1-point, 2-point, n-point, uniform).

- Mutation rate (0% - 10%).

- Elitist survival percentage (0% - 25%, tournament).

- Type of penalty function and its weight (linear, quadratic).

➔ A higher penalty does not necessarily increase the number of feasible solutions.

## Dynamic penalty weight:

- Dynamic penalties based on the quality of the best solution so far.

- The less feasible the best solution, the higher the penalty weight, to force the population closer to a feasible region.

- Once the best solution is feasible, the weight drops to a fixed low value $v$ to encourage improvement of the solution quality.

- For each generation the new penalty weight is calculated as follows, where $q$ is the number of violated constraints by the top solution and $\alpha$ is a preset severity parameter:

$$g_{demand} = \begin{cases} \alpha \cdot q & \text{for } q > 0 \\ v & \text{for } q = 0 \end{cases}$$

➔ The quality and feasibility of solutions is improved by approx. 10%.

# Sub-Population, Special Crossover and Migration

## Idea:

- 'Building Block Hypothesis' is only valid for problems with no or low epistacity.

- **BUT** the Nurse Scheduling Problem is highly epistatic due to the penalty function approach.

- Epistacity is present if the sum of the fitness of the single elements is bigger than the total fitness of the string.

- No/low epistacity means that if all single elements of a solution are good, then so is the whole solution.

[Davidor Y, *Epistasis Variance: A Viewpoint on GA-Hardness*, in Foundations of Genetic Algorithms, p 23ff, 1991.]

## **New type of 'grade based' crossover:**

- Idea: Split up the problem to reduce grade-based epistacity.

- Based on observation that 1-point crossover solves sometimes better than uniform crossover.

- Sort string according to nurses' qualifications.

- New type of crossover: fixed point crossover on grade-boundaries to keep good substrings together.

- On its own it limits the search space. Therefore combine with uniform crossover.

# Introduction of sub-populations and separate fitness functions:

- Have a separate population for each grade and for each combination of grades.

- Each sub-population follows an appropriate fitness function.

- In higher level sub-populations, parents are drawn from the population itself and from appropriate lower level populations.

- Then theses parents perform a suitable grade-based crossover.

- Have one additional and larger 'main' population which draws its parents from all others and has the original objective as fitness function.

**Fitness function of sub-populations:**

- Population 1/2/3 only optimise cover and penalty for grade 1/2/3.

- Population 4, 5, 6 and 7 optimise cover and penalty for grade 1+2, 1+3, 2+3 and 1+2+3 respectively.

- 'Main' population 8 optimises cover for grade 1, 1+2, 1+2+3 and penalty for all grades.

**Crossover:**

- Populations 1-3 perform uniform crossover for maximum diversity.

- Populations 4-7 perform 50% uniform and 50% grade-based crossover. In the case of grade-based crossover the parents are picked from niches 1-3 and combined accordingly.

- Population 8 performs 50% uniform and 50% grade-based crossover. In the case of grade-based crossover the parents are picked from niches 1-7 and combined accordingly.

**Results:**

- Feasibility and solution quality more than doubled.

- Loss of information due to limited choice.

➔ The later can be eradicated by the use of *migration*, i.e. by allowing individuals to change sub-populations over time.

# Incentives and Local Search/Repair

**Idea:**

- Remaining problems due to a day/night- shift imbalance.

- Once the wrong nurses are on days/nights the GA cannot sort it out.

- Look ahead and reward solutions with a 'future' potential, i.e. balanced solutions.

**Balanced Solutions:**

- A balanced solution has either both a day surplus *and* a day shortage *or* both a night surplus *and* a night shortage of nurses.

- An unbalanced solution has either both a day surplus *and* a night shortage **or** vice versa.

**Examples:**

|             | Days                    | Nights                  |
|-------------|-------------------------|-------------------------|
| Balanced:   | [+2  0  -1  0  -1  0  0] | [0  0  0  0  0  0  0]   |
| Unbalanced: | [0  0  +1  0  0  0  0]  | [0  -1  0  0  0  0  0]  |
| Neither:    | [0  -1  -1  +1  0  0  -2] | [0  0  +2  0  +2  0  -1] |

### Incentives/Disincentives:

- Balanced solutions receive a bonus towards their objective function value (incentive).

- Unbalanced solutions receive a negative bonus (disincentive).

➔ Balanced but less fit solutions rank higher than unbalanced solutions.

### Local Search/Repair:

- Take a balanced solution and use a first fit descending heuristic by cycling through each nurses' shift patterns.

- Better solutions are immediately accepted and the search is continued from there.

### Results:

- Disincentives immediately improve solution quality.

- Incentives improve solution quality when used with local search/repair.

# Delta Coding and Swaps

### **Delta Coding:**

- Idea: Seed the genetic algorithm with a population from within a hypercube around a previous good solution.

[Whitley D, *Delta Coding: An Iterative Search Strategy for Genetic Algorithms*, in Foundations of Genetic Algorithms 2 59ff, 1993.]

➔ Does not work for a discontinuous solution space.

### **Swaps:**

- Allow to swap worked shifts between nurses with the same qualification and working hours, as long as the sum of all their satisfied requests improves.

➔ Swaps are time expensive but offer a slight improvement in solution quality.

# Delta Coding

### Idea:

- Start the genetic algorithm with a population based on a previous good solution.

[Whitley D, *Delta Coding: An Iterative Search Strategy for Genetic Algorithms*, in Foundations of Genetic Algorithms 2 59ff, 1993.]

### Features:

- Run a standard genetic algorithm with a random starting population until finished.

- Run genetic algorithm again, but initialise new population as a hypercube surrounding the best previous solution.

- Due to reduced range of the variables the population size can be reduced, i.e. speed-up.

- Repeat until a local optimum is found.

### Results:

- Difficult to determine size of hypercube and whether a (local) optima is found.

- Problems with discrete variables.

- Does not work for a discontinuous solution space.

# Swaps

### **<u>Shift Swaps:</u>**

- Allow to swap worked shift patterns between all nurses with the same qualification and working hours, as long as the sum of all their satisfied requests improves.

- Swaps are time expensive and therefore restricted to the top solution in each generation.

### **<u>Special Swaps:</u>**

- Due to work contracts some nurses work either 3 day *or* 3 night shifts (3/3) whilst others work either 4 day *or* 3 night shifts (4/3).

- If a 4/3 nurse is on nights and a 3/3 nurse is on days the problem is unsolvable.

- The GA cannot resolve such a situation if a high degree of deceptiveness is present.

- A special swap operator is introduced to check if such a situation is present, and if so, it swaps the two nurses.

# Future Work

## Changes to the GA:

- Intelligent Mutation.

- Dynamic Parameters.

## Changes to the Coding:

- Employ an order based coding.

- Keep the GA simple.

- Use the problem specific information within the decoder.

- Concentrate on day/night problem?

## Apply the modified GA to other problems:

- Generalised assignment problems.

- Other multiple choice problems.

# Conclusions

- Basic GA cannot solve the problem.

- Stumbling block: epistacity creating constraints.

- Parameter/Strategy optimisation has little effect.

- Specialised operators are very effective.

- Local searches/repairs improve results further.

➔ Has the GA failed? No, because the problem is also difficult to solve with other methods (i.e. TABU search, IP programming).